\documentclass[10pt,twocolumn,letterpaper]{article}
\usepackage[pagenumbers]{cvpr}
\usepackage{amsmath,amssymb,bm}
\usepackage{graphicx,booktabs,multirow,array}
\usepackage{tikz}
\usetikzlibrary{arrows.meta,positioning,calc,fit}
\usepackage{pgfplots}
\pgfplotsset{compat=1.18}
\usepackage{microtype}
\usepackage{placeins}
\AtBeginDocument{\raggedbottom}
\definecolor{readoutblue}{RGB}{38,99,143}
\definecolor{controlorange}{RGB}{196,115,43}

\newcommand{\softplus}{\operatorname{softplus}}


\definecolor{cvprblue}{rgb}{0.21,0.49,0.74}
\usepackage[pagebackref,breaklinks,colorlinks,allcolors=cvprblue]{hyperref}
\usepackage[capitalize]{cleveref}
\def\confName{CVPR}
\def\confYear{2027}
\title{Fly as Instructed: Refining World–Action Policies for UAV Vision–Language Navigation}
\author{
{Zhexuan Zhou}\textsuperscript{1,*}\quad
{Yanke Sun}\textsuperscript{1,*}\quad
{Yuhan Zong}\textsuperscript{2}\quad
{Youmin Gong}\textsuperscript{1,\textdaggerdbl}\quad
{Jie Mei}\textsuperscript{3,\textdagger}\\[3pt]
\textsuperscript{1}Harbin Institute of Technology \quad
\textsuperscript{2}National University of Singapore\\
\textsuperscript{3}Harbin Institute of Technology, Shenzhen\\[3pt]
{\small\textsuperscript{*}These authors contributed equally.\quad
\textsuperscript{\textdagger}First corresponding author.\quad
\textsuperscript{\textdaggerdbl}Second corresponding author.}
}
\hypersetup{pdfauthor={Zhexuan Zhou, Yanke Sun, Yuhan Zong, Youmin Gong, Jie Mei}}
\begin{document}
\maketitle
\begin{abstract}
Natural language-guided UAV navigation enables intuitive human–robot interaction and flexible autonomous flight in complex environments. However, reliable navigation requires accurate instruction following, continuous task progress tracking, and safe obstacle avoidance, which remain challenging for existing world–action models that primarily focus on action prediction. To address these limitations, we propose a world–action learning framework that integrates three complementary mechanisms: a language-conditioned action interface with targeted post-training for instruction-responsive motion, recurrent instruction-consumption prediction to track task progress and determine termination, and Gaussian-scene post-training to enhance obstacle avoidance through trajectory imitation and differentiable geometric feedback. Closed-loop evaluations and ablation studies demonstrate improvements in sequential task completion, safe goal reaching, and timely termination. Furthermore, direct one-step denoising reduces inference overhead, and real-world indoor and outdoor flights validate the framework's effectiveness.
\end{abstract}

\section{Introduction}
\label{sec:intro}
Aerial vision--language navigation (VLN) enables unmanned aerial vehicles (UAVs) to follow natural-language instructions using visual observations~\cite{wang2025uavflow,Fan_2026_CVPR}. By expressing goals through objects, spatial relations and motion commands, users can specify flight tasks without constructing detailed waypoint sequences. This capability supports intuitive human--robot interaction and flexible autonomous navigation in complex environments. Reliable aerial VLN requires the vehicle to ground language in continuous three-dimensional motion, track the progress of multi-stage tasks, and avoid obstacles throughout execution.

Recent world--action models couple visual representations with action generation~\cite{yuan2026fastwam}, providing a foundation for aerial policies such as WorldVLN and WorldFly~\cite{worldvln2026,worldfly2026}. Turning these predictions into complete flights brings three learning challenges. First, action policies can rely on visual shortcuts when the requested motion changes~\cite{fang2026counterfactual}, making language-responsive control essential. Second, multi-stage instructions require persistent knowledge of completed requirements to guide subsequent actions and determine when to stop~\cite{ma2019selfmonitor,huang2026aiep}. Third, obstacle avoidance requires feedback on the path induced by predicted actions and the vehicle's motion response~\cite{zhang2025diffphys}. These challenges connect instruction grounding, execution history and physical motion within the same control loop.

In this paper, we propose a world--action UAV framework that connects language response, task progress and obstacle-aware motion (\cref{fig:final_pipeline}). The action expert reads the instruction through text cross-attention and undergoes targeted post-training for instruction-responsive motion while preserving pretrained conditioning modules. A recurrent instruction-consumption head predicts completed requirements, continuously reweights the complete instruction inside the frozen executor, and supplies task-level stopping. Safety post-training in reconstructed Gaussian scenes combines successful-detour imitation with geometric feedback differentiated through vehicle tracking dynamics. Constrained adapter updates retain general-task behavior, and direct one-step denoising reduces deployment cost.

A unified closed-loop evaluation and targeted controls assess navigation, progress and safety. Equal-budget data controls test language-conditioned response, while six indoor and outdoor flights demonstrate the integrated system.

Our contributions are:
\begin{itemize}
\item A language-conditioned action interface and targeted post-training design for instruction-responsive UAV motion, evaluated with matched training and paired text interventions.
\item A recurrent instruction-consumption interface that predicts completed requirements from visual and execution history, continuously reweights the complete instruction inside a frozen executor, and supplies task-level stopping.
\item Gaussian-scene obstacle-avoidance post-training that learns from successful detours and differentiable pathwise geometric feedback, with constrained updates that retain general-task behavior.
\end{itemize}

\section{Related Work}
\label{sec:related}
Aerial VLN draws on three lines of research: language-guided navigation, task progress estimation, and learning to avoid obstacles in flight.

\paragraph{Language-guided UAV navigation.}
VLN has developed from panoramic navigation graphs~\cite{anderson2018vln} to continuous environments~\cite{krantz2020vlnce} and long-horizon instruction execution~\cite{Song_2025_CVPR}. Aerial VLN extends this setting to three-dimensional motion: HTNav organizes aerial decisions hierarchically~\cite{Fan_2026_CVPR}, UAV-Flow provides fine-grained flight demonstrations and simulation evaluation~\cite{wang2025uavflow}, and FLIGHT supplies multi-stage demonstrations with state-reasoning text~\cite{zheng2026flight}. HUGE-Bench emphasizes process correctness and safety~\cite{guo2026hugebench}.

Generative visual models support trajectory planning and language grounding through visual imagination~\cite{Bar_2025_CVPR,Perincherry_2025_CVPR}. With pretrained video representations such as Wan~\cite{wan2025wan}, world--action models connect visual prediction and action generation~\cite{yuan2026fastwam}, with aerial examples including WorldVLN and WorldFly~\cite{worldvln2026,worldfly2026}. UAV-Flow++ further introduces online post-training for FlowVLA~\cite{yang2026uavflowplus}. CAST shows how counterfactual labels expand instruction--action diversity~\cite{glossop2025cast}, and LIBERO-CF shows how visual shortcuts can persist when instructions change~\cite{fang2026counterfactual}. We combine language conditioning with targeted action-expert post-training and use matched training and paired instructions to assess the resulting language response.

Action policies generate commands through diffusion, flow matching, or discrete tokens~\cite{chi2023diffusion,lipman2023flow,kim2024openvla}. Prior efficiency work uses visual compression and latent subgoals with consistency flow matching~\cite{Yang_2026_CVPR,Chen_2026_CVPR}. Our action expert supports direct one-step denoising, reducing deployment cost within the same policy architecture.

\paragraph{Task progress and instruction execution.}
Self-Monitoring learns navigation progress~\cite{ma2019selfmonitor}, and sub-instruction-aware navigation uses fine-grained path annotations to shift attention across instruction segments~\cite{hong2020subinstruction}. VLN-AIEP suppresses completed content using progress learned from navigation labels~\cite{huang2026aiep}, while Route2Step uses a step-level interface to connect semantic progress with local actions~\cite{huang2026route2step}. Coverage attention provides a related mechanism for tracking consumed content in sequence generation~\cite{tu2016coverage}. Our recurrent head learns token satisfaction from clause completion. The same completion state continuously reweights the full instruction inside a frozen action expert and determines when the UAV should stop.

\paragraph{Learning obstacle avoidance for UAVs.}
Gaussian splatting supports scene rendering~\cite{kerbl2023gaussian} as well as online planning and localization in Splat-Nav~\cite{chen2025splatnav} and action filtering in SAFER-Splat~\cite{chen2025safersplat}. Differentiable flight training shapes the policy itself by propagating feedback through simulated vehicle motion~\cite{zhang2025diffphys}. We use reconstructed Gaussian scenes to post-train a language-conditioned flight policy. The action expert learns from successful detours and geometric feedback differentiated through vehicle tracking dynamics.

Our world--action policy brings these lines of work together to execute complete flight tasks from natural language, using learned progress to guide execution and stopping and safety post-training to shape flight behavior.

\section{Method}
\label{sec:method}
\begin{figure*}[t]\centering
\includegraphics[width=\textwidth]{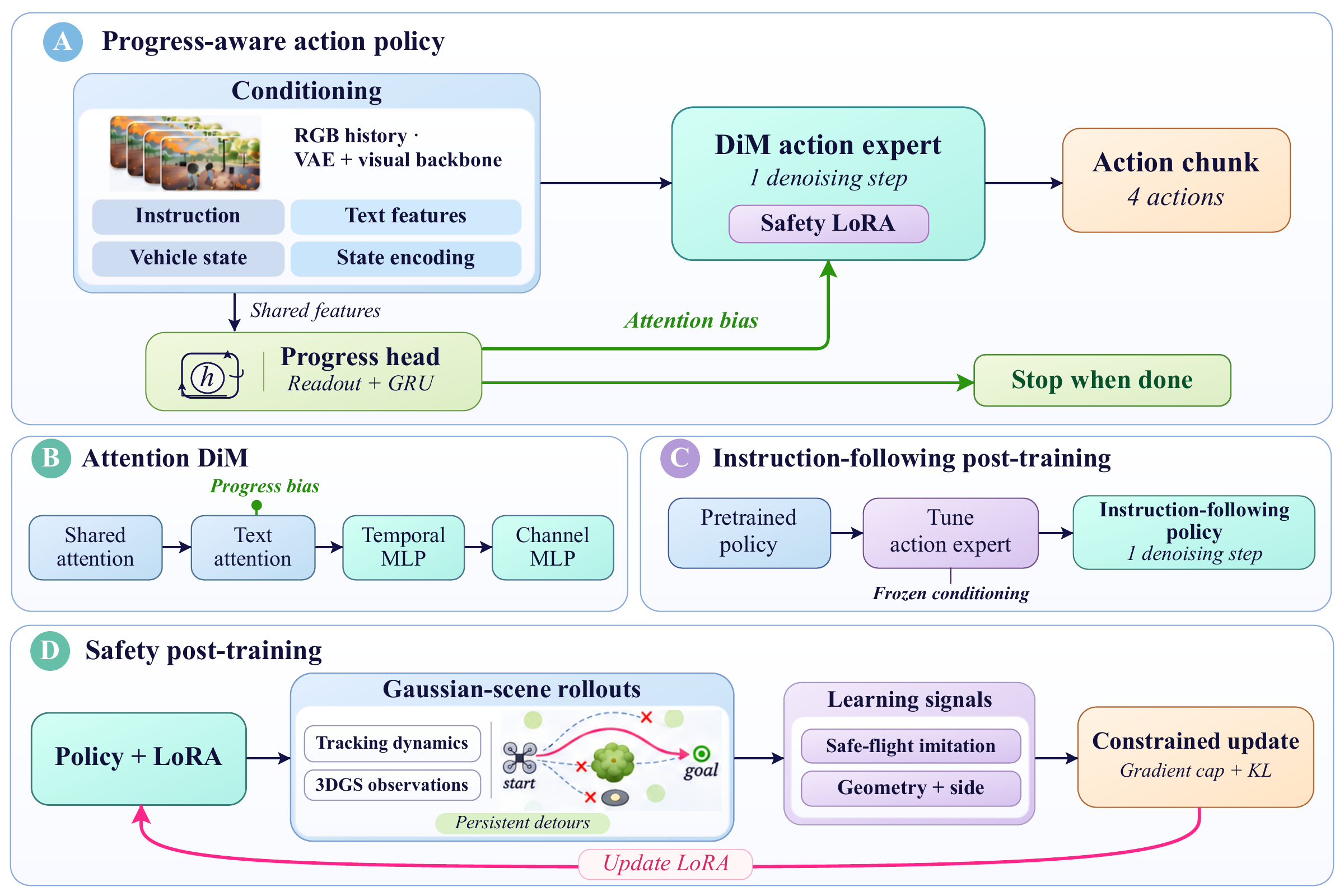}
\caption{\textbf{Progress-aware world--action policy and post-training.} (A) RGB history, the complete instruction and vehicle state condition a one-step action expert. The integrated recurrent progress head biases language attention and triggers stopping; safety adapters attach to the expert. (B) Attention DiM combines shared attention, text cross-attention, temporal mixing and channel mixing. (C) Targeted action-expert post-training produces an instruction-following policy deployed with one denoising step. (D) Gaussian-scene safety adaptation combines successful-flight imitation and geometric feedback through tracking dynamics, with constrained adapter updates. Rendering supplies observations outside the gradient path. Scene thumbnails are schematic.}
\label{fig:final_pipeline}\label{fig:execution_modules}\end{figure*}

The \emph{Action Policy} uses direct one-step denoising after post-training and supports two execution modules: a consumption head that reweights the complete instruction and triggers stopping, and safety adapters that shape obstacle-aware motion. The \emph{Full System} enables both. \Cref{fig:final_pipeline} summarizes their backbone and learning signals.

\subsection{Language-conditioned action learning}
At decision time $t$, the policy receives four RGB history frames $H_t$, proprioceptive state $s_t$ and instruction features $c$. It generates a four-action chunk of body-frame XYZ and yaw increments with one denoising step, using cached instruction features. A Wan-based video branch~\cite{wan2025wan} supplies visual conditioning to a 12-layer action expert; inference operates on the observed history without generating an explicit future video.

The action expert retains shared attention and text cross-attention, followed by an action mixer (DiM) with separate temporal and channel MLPs~\cite{tolstikhin2021mlpmixer}. Text cross-attention supplies language throughout action generation; the progress interface in \cref{sec:instruction_consumption} modulates its logits as requirements are completed. Implementation details are given in the supplement.

Starting from a UAV-pretrained policy, we post-train the action expert for instruction-responsive motion while freezing the visual, VAE and proprioceptive modules. The resulting policy directly supports one-step denoising for deployment; a UAV-only continuation provides an equal-budget data control. Training schedules, data composition and matched evaluation settings are detailed in the supplement.

\subsection{Instruction consumption for progress-aware execution}
\label{sec:instruction_consumption}
The complete instruction stays in context throughout execution. Recurrent token satisfaction continuously shifts attention toward unfinished requirements (\cref{fig:final_pipeline}A).

\paragraph{Instruction progress prediction.}
For instruction tokens $w_1,\ldots,w_N$, a recurrent head predicts satisfaction probabilities $m_{t,i}$. Let $e_i$ be a projected T5~\cite{raffel2020t5} embedding and $v_i$ mark task-bearing tokens; the ordinary text mask excludes padding separately. The remaining-instruction query is
\begin{equation}
r_t=\frac{\sum_i v_i(1-m_{t-1,i})e_i}
{\max(\sum_i v_i(1-m_{t-1,i}),\varepsilon)}.
\end{equation}
It attends over the frozen backbone features $V_t$ that the executor already computes. A GRU combines the attended feature, $r_t$, ego-state encoding $s_t$ and edit flag $b_t$:
\begin{align}
h_t&=\operatorname{GRU}(h_{t-1},f(\operatorname{Attn}(r_t,V_t),r_t,s_t,b_t)),\\
m_{t,i}&=\operatorname{sigmoid}(g([h_t,e_i])).
\end{align}
The recurrent memory spans the task, and the head reuses the policy's visual and language representations. State features encode heading and causal time.

FLIGHT's clause-completion labels~\cite{zheng2026flight} supervise token satisfaction through clause-balanced binary cross-entropy. Deployment feeds predicted satisfaction back into the next decision. The supplement details supervision, state encoding and instruction-edit handling.

\paragraph{Progress-guided action generation.}
The soft-consumption mask acts where the executor reads language. Let $\ell_{k,i}$ be the context cross-attention logit from action position $k$ to language token $i$. We add a token-wise bias,
\begin{equation}
\tilde\ell_{k,i}=\ell_{k,i}+\beta\,v_i\log\!\big(1-m_{t,i}+\varepsilon\big),
\label{eq:consumption_bias}
\end{equation}
in each action-expert block, with $\beta>0$ and $\varepsilon>0$. Increasing satisfaction lowers a task token's logit; softmax jointly renormalizes all attention weights. Other positions retain their logits. Soft Mask preserves the complete text. Hard Switching uses the same completion representation but re-encodes the earliest unfinished clause, changing both context and conditioning.

The same satisfaction estimates determine when to stop. For clause $C_a$, let $\bar m_{t,a}=|C_a|^{-1}\sum_{i\in C_a}m_{t,i}$. A clause is latched complete when $\bar m_{t,a}>\tau$ for $k_s$ consecutive decisions. Completion of every clause issues the stop command under either conditioning interface. We select $\tau$ and $k_s$ on development flights and freeze them for evaluation. As the controller executes each action chunk, new observations update satisfaction and guide the next prediction.

\subsection{Gaussian-scene safety adaptation}
\label{sec:gaussian_action_posttraining}
We train low-rank action-expert adapters~\cite{hu2022lora} in reconstructed Gaussian scenes~\cite{kerbl2023gaussian}, freezing the remaining policy. A bounded tracking response executes body-frame displacement and yaw commands with velocity, acceleration and delay. The same map $F$ advances safety rollouts and supplies geometric gradients with observations fixed. Gaussian rendering supplies observations outside the gradient path. The supplement gives the dynamics and objectives.

\paragraph{Persistent exploration and outcome filtering.}
A sampled lateral detour is held over an initial window and released at a displacement cap. Paired rollouts share the scene, start pose and action noise, while mirrored detours explore both sides~\cite{raffin2022gsde,plappert2018parameter}. The full instruction remains available after the detour; progress updates change its attention bias. Successful contact-free flights with positive group-relative advantage enter replay. Self-imitation~\cite{oh2018selfimitation} regresses predicted actions toward detached successful executed actions, with replay balanced across scenes and directions.

\paragraph{Differentiable geometric feedback.}
Following differentiable flight training~\cite{zhang2025diffphys}, we roll the predicted action chunk through the motion map $F$ to obtain a path $Q_\theta=F(q_t,\operatorname{denorm}(A_t))$. Registered geometry supplies swept-segment clearance $\tilde d_j$, including vehicle radius and segment sampling. For margin $m$, smoothness $\epsilon$ and prefix length $k$,
\begin{equation}
\mathcal L_{\rm clear}=\frac1k\sum_{j=1}^k
\left[\epsilon\softplus\!\left(\frac{m-\tilde d_j}{\epsilon}\right)\right]^2,
\end{equation}
The geometric loss adds a workspace penalty and navigation potential: $\mathcal L_{\rm geo}=\mathcal L_{\rm clear}+\mathcal L_{\rm wall}+\Phi(p_k)-\Phi(p_0)$. Passage-side feedback contributes $\mathcal L_{\rm side}$. These auxiliary terms shape training; evaluation uses visual and instruction conditioning without exploration detours or obstacle-geometry input.

\paragraph{Constrained adaptation.}
Let $g_{\rm SI}$ be the imitation gradient and $g_{\rm aux}=\nabla_\theta(\mathcal L_{\rm geo}+\lambda_{\rm s}\mathcal L_{\rm side})$. We cap the auxiliary contribution by the imitation-gradient norm,
\begin{equation}
g=g_{\rm SI}+\eta\,c_g\,g_{\rm aux},\quad
\eta=\min\!\left(1,\frac{\|g_{\rm SI}\|}{\max(c_g\|g_{\rm aux}\|,\varepsilon)}\right),
\end{equation}
and skip updates with zero imitation gradient, so adaptation requires successful replay. A KL trust region~\cite{schulman2015trust} constrains the action change on rollout contexts, using a fixed-covariance Gaussian surrogate around the action mean. The updated policy collects the next rollouts. Uncertain contacts are excluded from training losses and retained in evaluation denominators. The ablations remove persistence, filtering, the auxiliary-loss bundle, or the norm/KL-constraint bundle.

\begin{table*}[!t]\centering\small
\setlength{\tabcolsep}{3pt}\renewcommand{\arraystretch}{1.12}
\caption{\textbf{Overall results on the unified test suite.} All methods are evaluated on 521 episodes. Entries report successful episodes (percentage), except nDTW, which lies in $[0,1]$.}
\label{tab:unified_test_results}
\begin{tabular*}{\textwidth}{@{\extracolsep{\fill}}lrrrrrr@{}}\toprule
Method & $\mathrm{SR}_{\rm reach}\uparrow$ & $\mathrm{SR}_{\rm stop}\uparrow$ & nDTW $\uparrow$ & OC $\uparrow$ & SFR $\uparrow$ & CFR $\uparrow$\\\midrule
\textbf{Ours (Full System)} & \textbf{436 (83.7)} & \textbf{401 (77.0)} & \textbf{0.648} & \textbf{416 (79.8)} & \textbf{399 (76.6)} & \textbf{452 (86.8)}\\
Ours (Action Policy) & 284 (54.5) & 0 (0.0) & 0.503 & 227 (43.6) & 210 (40.3) & 318 (61.0)\\
WorldVLN & 236 (45.3) & 164 (31.5) & 0.448 & 181 (34.7) & 158 (30.3) & 305 (58.5)\\
FastWAM & 221 (42.4) & 0 (0.0) & 0.431 & 168 (32.2) & 149 (28.6) & 287 (55.1)\\
HugeBench & 207 (39.7) & 139 (26.7) & 0.417 & 155 (29.8) & 137 (26.3) & 293 (56.2)\\
Pi0-UAV & 178 (34.2) & 97 (18.6) & 0.386 & 131 (25.1) & 118 (22.6) & 301 (57.8)\\
OpenVLA-UAV & 143 (27.4) & 88 (16.9) & 0.352 & 102 (19.6) & 91 (17.5) & 279 (53.6)\\
FlowUAV & 119 (22.8) & 61 (11.7) & 0.307 & 83 (15.9) & 72 (13.8) & 296 (56.8)\\
\bottomrule\end{tabular*}
\par\vspace{4pt}\begin{minipage}{\textwidth}\footnotesize SFR measures safe-flight success: reaching the task goal with contact-free execution. CFR is the collision-free rate over all episodes. Reach, stop, OC, SFR and CFR share the denominator 521; multi-stage completion follows the ordered task requirements in \cref{sec:protocol}.\end{minipage}
\end{table*}

\begin{table*}[!t]\centering\small
\setlength{\tabcolsep}{3pt}\renewcommand{\arraystretch}{1.12}
\caption{\textbf{Module independence.} The same action policy is evaluated with the safety adapter, the instruction-consumption head, and both modules. Entries report successful episodes (percentage).}
\label{tab:module_independence}
\begin{tabular*}{\textwidth}{@{\extracolsep{\fill}}lrrrr@{}}\toprule
Configuration & Safety SFR & Progress OC & Progress $\mathrm{SR}_{\rm stop}$ & General $\mathrm{SR}_{\rm reach}$\\\midrule
Action Policy & 52 (30.6) & 41 (24.1) & 0 (0.0) & 139 (76.8)\\
+ Safety Adapter & 133 (78.2) & 42 (24.7) & 0 (0.0) & 138 (76.2)\\
+ Consumption Head & 54 (31.8) & 138 (81.2) & 131 (77.1) & 151 (83.4)\\
\textbf{+ Both (Full System)} & \textbf{133 (78.2)} & \textbf{138 (81.2)} & \textbf{131 (77.1)} & \textbf{152 (84.0)}\\
\bottomrule\end{tabular*}
\par\vspace{4pt}\begin{minipage}{\textwidth}\footnotesize The safety and progress subsets each contain 170 episodes; the general-task subset contains 181 episodes. These subsets form the unified 521-episode test suite.\end{minipage}
\end{table*}

\section{Experimental Setup}
\label{sec:protocol}
\label[section]{tab:configurations}
\paragraph{Data and tasks.}
We evaluate 521 closed-loop episodes: 181 general-task, 170 safety, and 170 progress episodes, covering UAV-Flow subtasks, composite instructions and complex scenes. Test instances or layouts differ from training configurations. Main comparisons use the complete suite; module ablations use the corresponding subsets.

\paragraph{Comparison methods.}
We compare reference systems from OpenVLA~\cite{kim2024openvla}, $\pi0.5$~\cite{black2025pi05}, UAV-Flow++~\cite{yang2026uavflowplus}, HUGE-Bench~\cite{guo2026hugebench}, WorldVLN~\cite{worldvln2026}, and Fast-WAM~\cite{yuan2026fastwam}. Methods share evaluation tasks, initial states, instructions, budgets and scoring rules while retaining native model interfaces. Internal controls cover equal-budget post-training and the progress and safety modules.

\paragraph{Implementation.}
Post-training updates the action expert with frozen conditioning modules. Latency is measured on an RTX 5880 Ada at batch one using four $256\times256$ history frames, four output actions and cached text. Full-model timing covers observation encoding, action generation and output transfer; HTTP timing is reported separately. Six real-world tasks each have one recorded flight using the final teacher with direct one-step denoising, progress consumption and safety adaptation on an RTX 5070, with 10\,Hz trajectory updates and 50\,Hz control.

\paragraph{Metrics.}
We report goal reach ($\mathrm{SR}_{\rm reach}$), correct stopping ($\mathrm{SR}_{\rm stop}$), path fidelity (nDTW~\cite{ilharco2019ndtw}), ordered completion (OC), safe-flight success (SFR), and collision-free rate (CFR). Correct stopping requires ordered completion and policy-issued termination. Failed and uncertain outcomes remain in the denominators. Task retention measures reach success on the 181 general tasks. The supplement specifies scoring events, baseline configurations, training parameters and auxiliary diagnostic protocols.

\section{Results}
\label{sec:results}
We evaluate complete flights, the contribution of each learning mechanism, and efficient physical deployment.

\subsection{Overall results on the unified test suite}
\label{sec:unified_test_results}
The full system improves arrival, ordered execution, and termination together. \Cref{tab:unified_test_results} reports aggregate performance on the unified 521-episode test suite. Ours (Full System) achieves 83.7\% goal-reach success, 77.0\% stop success, 0.648 nDTW, 79.8\% OC, and 76.6\% SFR, with an 86.8\% collision-free rate. WorldVLN is the strongest external reference on reach, stop, nDTW, OC and SFR, scoring 45.3\%, 31.5\%, 0.448, 34.7\% and 30.3\%, respectively. The full system improves reach by 38.4 percentage points and stop success by 45.5 points over this reference.

Relative to our action policy, adding safety adaptation and instruction consumption raises reach from 54.5\% to 83.7\%, OC from 43.6\% to 79.8\%, and SFR from 40.3\% to 76.6\%.

\paragraph{Complementary module effects.}
\Cref{tab:module_independence} separates the module effects. Safety adaptation raises safety-subset SFR from 30.6\% to 78.2\%, with progress OC changing from 24.1\% to 24.7\%. Consumption yields 81.2\% progress OC and 77.1\% stop success. Together, the modules retain 78.2\% safety SFR, 81.2\% progress OC, and 77.1\% progress stop success, while general-task reach rises from 76.8\% to 84.0\%. The combination retains the reported gains in safe arrival, ordered execution, and stopping across these subsets.

\begin{figure*}[!t]
\centering
\includegraphics[width=\textwidth]{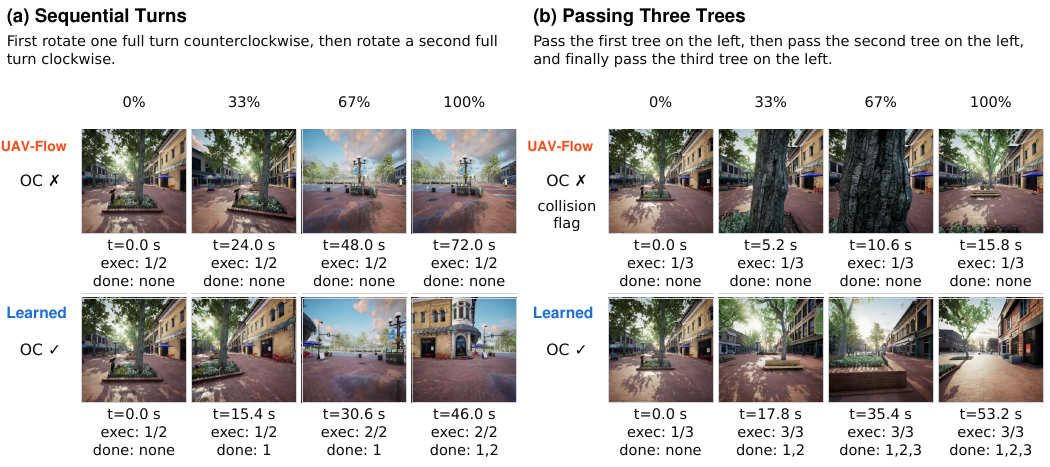}
\caption{\textbf{Task-progress prediction and execution on matched tasks.} Upper rows: UAV-Flow, reproduced following its official method as in the main-table evaluation. Lower rows (Learned): our method, which retains the full instruction and attenuates attention to completed tokens using predicted progress. (a) A full counterclockwise turn followed by a full clockwise turn. (b) Passing three trees in sequence on the left. Columns show 0/33/67/100\% of each rollout's duration. Stage indices and \texttt{exec}/\texttt{done} annotations visualize ground-truth task progress; our online attention weights are determined by the progress predictor. OC denotes ordered completion.}
\label{fig:main_progress_comparison}
\end{figure*}

\subsection{Language-instruction following}
\Cref{tab:main_text_controls} compares our method with Pi0-UAV, WorldVLN, and OpenVLA-UAV on instruction execution, sensitivity to semantic changes, and robustness to paraphrasing. In the same UE simulation environment, Instruction SR uses 500 individual instructions. Contrastive SR and Paraphrase SR each use 500 instruction pairs. Instruction SR requires satisfying all applicable trajectory and terminal constraints. Contrastive SR requires both instructions in a semantic contrast pair to succeed from the same initial state; Paraphrase SR applies the same criterion to equivalent expressions.

Ours achieves 80.6\% Instruction SR, 56.4\% Contrastive SR, and 38.2\% Paraphrase SR, leading all three metrics. Instruction SR exceeds WorldVLN, the strongest baseline on this metric, by 38.4 percentage points. On the paired metrics, Ours exceeds Pi0-UAV by 35.6 points in Contrastive SR and 29.0 points in Paraphrase SR. These results demonstrate more reliable instruction execution, stronger sensitivity to changes in task meaning, and greater robustness to equivalent wording across the evaluated tasks.

\begin{table}[!htbp]\centering\small
\setlength{\tabcolsep}{2pt}\renewcommand{\arraystretch}{1.15}
\caption{\textbf{Language-instruction-following comparison.} Success rates (\%) over 500 individual instructions for Instruction SR and 500 instruction pairs for each paired metric. A pair succeeds only if both instructions succeed.}
\label{tab:main_text_controls}
\begin{tabular*}{\columnwidth}{@{\extracolsep{\fill}}>{\raggedright\arraybackslash}p{0.32\columnwidth}ccc@{}}\toprule
Method & \shortstack{Instruction\\SR (\%) $\uparrow$} & \shortstack{Contrastive\\SR (\%) $\uparrow$} & \shortstack{Paraphrase\\SR (\%) $\uparrow$}\\\midrule
Pi0-UAV & 33.2 & 20.8 & 9.2\\
WorldVLN & 42.2 & 18.4 & 5.2\\
OpenVLA-UAV & 26.4 & 13.6 & 8.2\\
\textbf{Ours} &  \textbf{80.6} &  \textbf{56.4} &  \textbf{38.2}\\\bottomrule
\end{tabular*}
\end{table}

\subsection{Clause execution and learned task progress}
\label{sec:progress_results_examples}
\paragraph{Instruction-consumption ablation.}
\Cref{tab:progress_control_ablation} compares five configurations on 170 progress episodes. Soft masking achieves 81.2\% OC and 77.1\% stop success, compared with 70.0\% and 61.2\% for hard switching, and 52.4\% and 47.6\% for timed switching. Removing GRU memory reduces OC to 57.6\% and stop success to 49.4\%, corresponding to drops of 23.5 and 27.6 percentage points. The full-instruction control records 24.1\% OC and 0.0\% stop success. Soft Mask improves over Hard Switching with the same progress head while retaining full-text context. This compares conditioning interfaces; Full Instruction also lacks the head's stop capability. The GRU ablation measures execution-memory contribution.

\begin{table}[!htbp]\centering\small
\setlength{\tabcolsep}{3pt}\renewcommand{\arraystretch}{1.12}
\caption{\textbf{Instruction-consumption ablation.} Each configuration is evaluated on 170 progress episodes. Entries report successful episodes (percentage).}
\label{tab:progress_control_ablation}
\begin{tabular*}{\columnwidth}{@{\extracolsep{\fill}}lrr@{}}\toprule
Configuration & OC $\uparrow$ & $\mathrm{SR}_{\rm stop}\uparrow$\\\midrule
Full Instruction & 41 (24.1) & 0 (0.0)\\
Timed Switching & 89 (52.4) & 81 (47.6)\\
Hard Switching & 119 (70.0) & 104 (61.2)\\
w/o GRU Memory & 98 (57.6) & 84 (49.4)\\
\textbf{Soft Mask (Ours)} & \textbf{138 (81.2)} & \textbf{131 (77.1)}\\
\bottomrule\end{tabular*}

\end{table}

\paragraph{Task-progress prediction and execution.}
\Cref{fig:main_progress_comparison} compares UAV-Flow and our method on two sequential tasks: consecutive turns in opposite directions and passing three trees in order. In these examples, UAV-Flow remains at the first task stage, whereas our method completes the ordered requirements. Our progress predictor guides execution by reducing attention to completed instruction tokens while preserving the full instruction context. The stage indices and completion labels provide ground-truth annotations for visualization. Online attention reweighting uses predicted progress. Frames are sampled at relative positions along each method's own rollout.

\begin{figure*}[t]
\centering
\includegraphics[width=\textwidth]{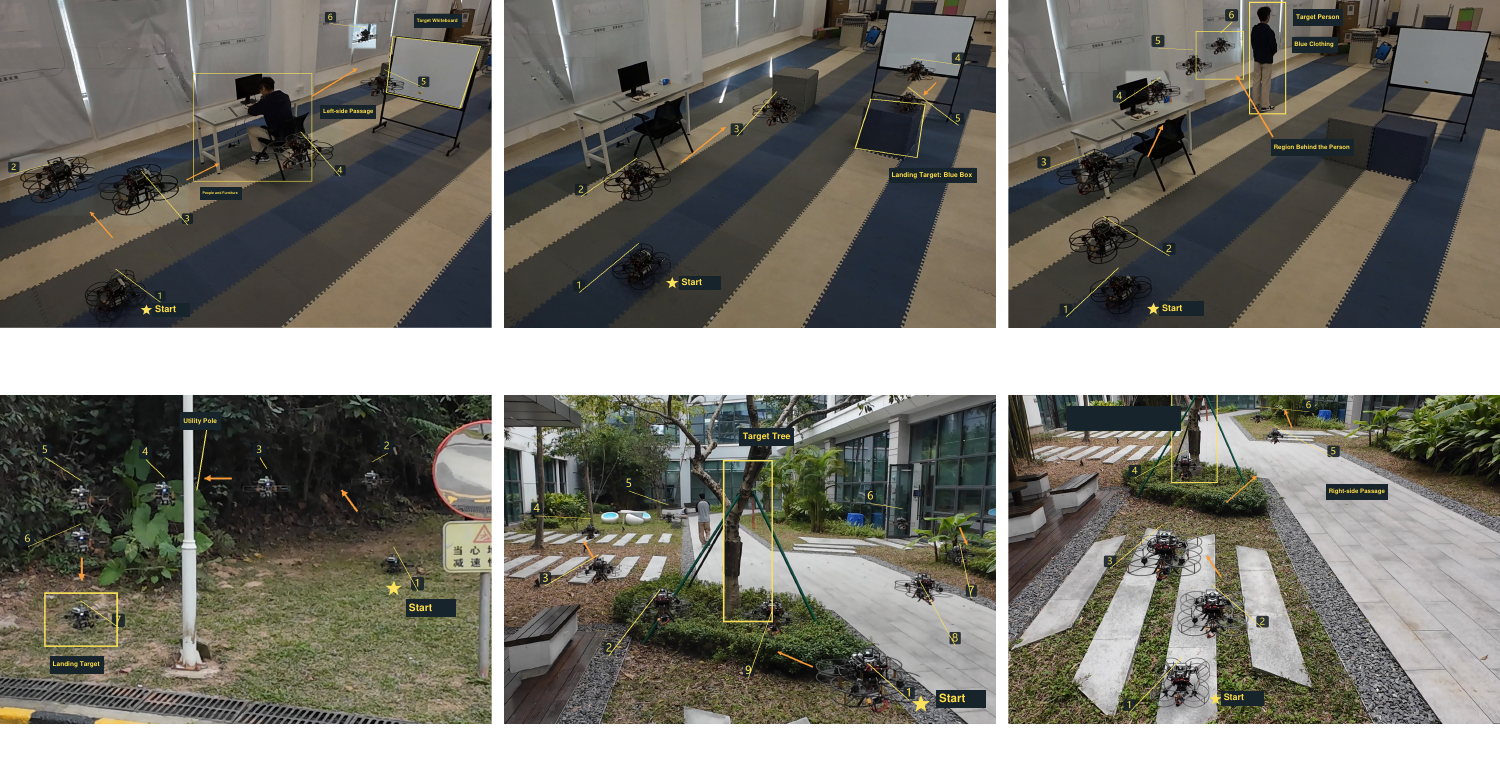}
\caption{\textbf{Real-world execution of the proposed policy.} All six tasks: (a) indoor obstacle passage, (b) box landing, (c) reaching behind a person, (d) pole-side landing, (e) tree circumnavigation and (f) right-side tree passage. Each panel overlays registered keyframes from one flight; numbered UAV positions indicate temporal order. Orange arrows indicate image-plane sequence order. The supplement provides enlarged sequences and platform details.}
\label{fig:real_world_overview}
\end{figure*}

\subsection{Targeted safety-navigation ablation}
\label{sec:avoidance_results}
\paragraph{Safety-adaptation ablation.}
\Cref{tab:safety_control_ablation} evaluates the safety adapter on 170 safety episodes and measures task retention on 181 general-task episodes. The full method achieves 78.2\% SFR with a 17.1\% collision rate, compared with 30.6\% and 64.1\% for the base policy. Replacing persistent exploration with per-step noise reduces SFR to 46.5\%; removing outcome filtering reduces it to 40.0\%. Removing auxiliary signals yields 59.4\% SFR. Removing constrained updates retains 72.9\% SFR but lowers general-task reach to 53.6\%, compared with 76.2\% for the full method and 76.8\% for the base policy. The auxiliary ablation measures geometric and side feedback jointly; the update ablation measures the norm cap and KL constraint jointly. These bundles improve safe arrival and preserve general-task behavior.

\begin{table}[!htbp]\centering\small
\setlength{\tabcolsep}{3pt}\renewcommand{\arraystretch}{1.12}
\caption{\textbf{Safety-adaptation ablation.} SFR and collision rate use 170 safety episodes. Task retention is goal-reach success on the 181 general-task episodes. Entries are counts (percentages).}
\label{tab:safety_control_ablation}
\begin{tabular*}{\columnwidth}{@{\extracolsep{\fill}}>{\raggedright\arraybackslash}p{0.39\columnwidth}rrr@{}}\toprule
Configuration & SFR $\uparrow$ & \shortstack{Collision\\rate $\downarrow$} & \shortstack{Task\\retention $\uparrow$}\\\midrule
Base & 52 (30.6) & 109 (64.1) & 139 (76.8)\\
w/o Persistent Exploration & 79 (46.5) & 84 (49.4) & 134 (74.0)\\
w/o Outcome Filtering & 68 (40.0) & 88 (51.8) & 124 (68.5)\\
w/o Auxiliary Signals & 101 (59.4) & 60 (35.3) & 141 (77.9)\\
w/o Constrained Updates & 124 (72.9) & 33 (19.4) & 97 (53.6)\\
\textbf{Full Method} & \textbf{133 (78.2)} & \textbf{29 (17.1)} & 138 (76.2)\\
\bottomrule\end{tabular*}
\par\vspace{4pt}\begin{minipage}{\columnwidth}\footnotesize Persistent exploration holds the exploration perturbation across successive steps; its ablation uses independent per-step noise. The remaining ablations remove outcome filtering, auxiliary signals, or constrained updates, respectively.\end{minipage}
\end{table}

Supplementary first-person sequences show successful detours and a wrong-way failure of a fixed safety candidate.

\subsection{Inference efficiency}
Our model has the fewest parameters (815.28M) and lowest median action-head and full-model latencies (30.55 and 91.84\,ms) in \cref{tab:main_efficiency}. Compared with FastWAM, it uses 54.8\% fewer parameters, runs its action head $2.75\times$ faster, and reduces full-model latency by 7.9\%. Full-model inference is $2.29\times$ faster than OpenVLA-UAV. Baselines use native configurations, and WorldVLN's 6167.84\,ms includes video generation. The supplement details timing boundaries and action-parameter counts.
\begin{table}[tbp]\centering\small
\setlength{\tabcolsep}{3pt}
\caption{\textbf{Inference comparison on RTX 5880 Ada.} Batch one, 100 timed calls; head/full P50 in ms, total parameters in millions. External rows use native configurations; WorldVLN includes video generation. The supplement gives action-parameter counts and timing details.}
\label{tab:main_efficiency}
\begin{tabular*}{\columnwidth}{@{\extracolsep{\fill}}lrrr@{}}\toprule
Model & Params (M) & Head (ms) & Full (ms)\\\midrule
FastWAM & 1802.59 & 84.13 & 99.67\\
OpenVLA-UAV & 7541.24 & 183.42 & 210.27\\
WorldVLN & 9181.06 & 302.51 & 6167.84\\\midrule
\textbf{Ours} & 815.28 & \textbf{30.55} & \textbf{91.84}\\\bottomrule
\end{tabular*}
\end{table}

\subsection{Real-world flight demonstrations}
\label{sec:real_world_results}
The six physical flights demonstrate object-relative motion and task execution in indoor and outdoor environments. Indoors, the UAV avoids a seated person and furniture before passing to the left of a whiteboard, approaches and lands on a blue box, and reaches the region behind a person wearing blue. Outdoors, it approaches the pole-side landing region and descends, completes a circuit around a tree, and passes a tree on its right. The pole keyframes show the final descent; the recordings demonstrate autonomous task execution. \Cref{fig:real_world_overview} shows all six tasks; the supplement provides larger views and time-stamped detail frames for each flight.


\section{Discussion and Limitations}
\label{sec:discussion}
High-speed flight leaves less time to react to new observations, motivating further reductions in end-to-end policy latency. The recurrent head tracks progress through an instruction, while much longer missions may also require recalling landmarks or decisions from earlier in the flight. Extending the policy with longer-term memory is a direction for these tasks. Generalization can also improve in environments whose layouts or visual conditions differ substantially from those seen during training.

\section{Conclusion}
We present a world--action policy for UAV vision--language navigation that combines targeted action post-training, recurrent task progress prediction, and obstacle avoidance learned in reconstructed Gaussian scenes. In simulation, the method leads the evaluated baselines in complete instruction execution and in joint success on contrastive and paraphrased instruction pairs. Recurrent progress estimates guide attention to unfinished requirements and support timely stopping. Safety post-training combines successful detour imitation with geometric feedback through a differentiable tracking model, improving safe goal reaching while preserving general navigation performance. The action policy generates actions in one denoising step, and six indoor and outdoor flights demonstrate deployment of the integrated system.

\end{document}